\documentclass[10pt,twocolumn,letterpaper]{article}

\usepackage{cvpr}
\usepackage{times}
\usepackage{epsfig}
\usepackage{graphicx}
\usepackage{amsmath}
\usepackage{amssymb}
\usepackage{xcolor}
\usepackage{verbatim}
\usepackage{caption}
\usepackage{enumitem}
\usepackage{appendix}

\usepackage[pagebackref=true,breaklinks=true,letterpaper=true,colorlinks,bookmarks=false]{hyperref}

\cvprfinalcopy 

\def\cvprPaperID{0000} 

\newcommand{\bt}{\mathbf{t}}

\newcommand{\bx}{\mathbf{x}}

\newcommand{\bE}{\mathbf{E}}

\newcommand{\bI}{\mathbf{I}}

\newcommand{\bO}{\mathbf{O}}

\newcommand{\bphi}{\mbox{\boldmath$\phi$}}

\newcommand{\btheta}{\mbox{\boldmath$\theta$}}

\DeclareMathOperator*{\argmax}{arg\,max}

\newcommand{\rev}[1]{#1}

\begin{document}

\title{Neural Contours: Learning to Draw Lines from 3D Shapes  \vspace{-5mm} }

\author{ 
        Difan Liu$^1$
    \and
        Mohamed Nabail$^1$
    \and
        Aaron Hertzmann$^2$
    \and
        Evangelos Kalogerakis$^1$
     \and \vspace{-4mm}
     \\ 
    $^1$University of Massachusetts Amherst \,\,\,\,\,\,\,\,\,\,\,\,\,\,\, $^2$Adobe Research
}

\twocolumn[{%
 \renewcommand\twocolumn[1][]{#1}%
 \maketitle
 \thispagestyle{empty}
 \vspace{-8mm}
 \centering
 \includegraphics[width=\textwidth]{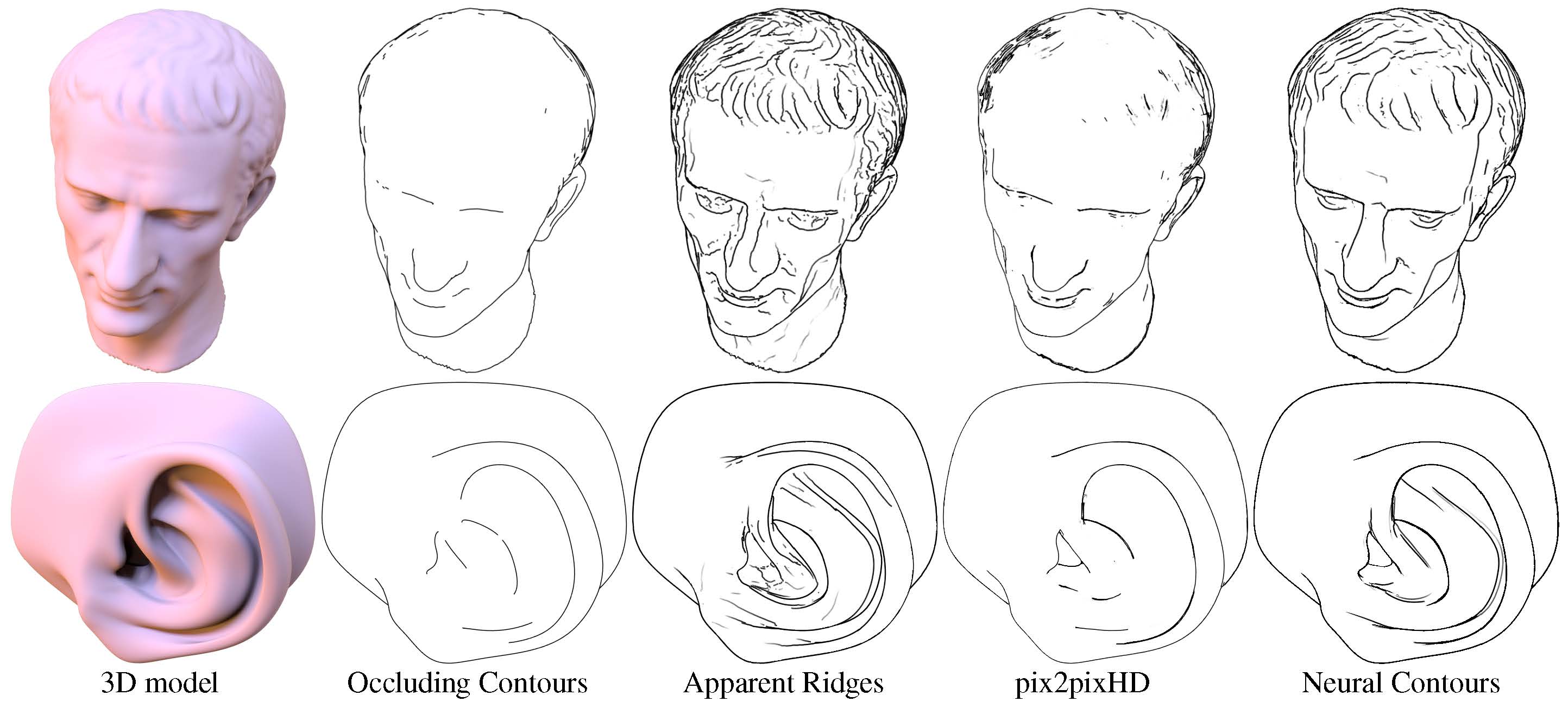}
 \vspace{-6mm}
 \captionof{figure}
  { 
    Given a 3D model (left), our network creates a line drawing that conveys its structure more accurately compared to using other methods individually, such as Occluding Contours \cite{BenardHertzmann}, Apparent Ridges \cite{Judd:2007}, and pix2pix variants  \cite{wang2018high}. Please see appendix for results on many more models.
    \label{fig:teaser}
   }
   
\vspace{3mm}   
}]

\begin{abstract}
\vspace{-3mm}       
    This paper introduces a method for learning to generate line drawings from 3D models. Our architecture incorporates a differentiable module operating on  geometric features of the 3D model, and an image-based module operating on   view-based shape representations. 
    At test time, geometric and view-based reasoning are combined with the help of a neural module to create a line drawing. The model is trained on a large number of crowdsourced comparisons of line drawings. Experiments demonstrate that our method achieves significant improvements in line drawing over the state-of-the-art when evaluated on standard benchmarks, resulting in drawings that are comparable to those produced by experienced human artists.   
\end{abstract}

\vspace{-4mm}
\section{Introduction}
 To draw an accurate line drawing of an object, an artist must understand the 3D shape, and select lines that will communicate that shape to a viewer. Modeling how artists perform this process remains an important open problem, intimately tied to 3D shape understanding, visual perception, and image stylization.  
Recent image stylization and image translation algorithms learn styles from examples, but do not take the underlying 3D geometry into account and generally do not capture outline drawing styles well.  In contrast, analytic geometry-based algorithms effectively capture basic line drawing properties, and have been used in many computer animation films and games. However, these methods can still fail to capture properties of real drawings, and require parameter tuning for each individual model.

This paper presents a method for learning to generate line drawings from 3D models. 
Our model employs two branches. The first branch implements geometric line drawing based on suggestive contours, apparent ridges, ridges, and valleys. Existing geometric line drawing approaches employ hard-to-tune user-specified parameters that need to be determined separately for each object.
In contrast, our method learns to automatically select these parameters through a differentiable module.  The second branch is represented as a standard image translation network, but using view-based shape representations as input. We show that combining both branches produces the best results.

A number of challenges exist in developing such an approach. First, classic geometric lines are not readily differentiable with respect to their parameters.
We combine soft thresholding functions along with image-space rendered maps of differential surface properties. Another major challenge is to combine the different geometric lines with purely learned ones. We show that a ranking module trained to assess the plausibility of the line drawing can be used to drive this combination. Finally, another challenge is to gather suitable training sets. Previous work has performed laborious in-person data collection from artists \cite{Cole:2008}. While this can provide high-quality datasets, it does not scale well, and it also produces data without much stylistic consistency. Can we train effective drawing models without employing artists to create many drawings? We describe a crowdsourcing approach to gather data using unskilled crowdworkers for ranking evaluations. 

We evaluated our method based on Cole \etal's \cite{Cole:2008} artist-created drawings, and find that our crowdsourced training produces state-of-the-art results.  We also gathered a new extended test dataset involving more diverse 3D models following  Cole \etal's methodology. In all test datasets, our method generated line drawings that are significantly more accurate and perceptually more similar to artists' drawings compared to prior work, including geometric methods and image translation networks.
We also conducted a MTurk evaluation study. We found that crowdworkers selected our line drawings as the best to convey a reference 3D model twice as frequently compared to the other techniques.

\vspace{-1mm}
\section{Related Work}
 How do artists create line drawings, and how do people perceive them? This question has been studied in art history \cite{Gombrich}, philosophy \cite{Goodman}, neuroscience \cite{SayimCavanagh}, and perceptual psychology \cite{Hertzmann:2020:WDL,Kennedy1974, Koenderink:1982}. Occluding contour algorithms are the foundation of non-photorealistic 3D computer graphics; see \cite{BenardHertzmann} for a survey.

Generalizations of occluding contours improved line drawing algorithms, beginning with the suggestive contours \cite{DeCarlo:2003, Lee:2007}, and continuing with Apparent Ridges \cite{Judd:2007} and several other methods; see \cite{DeCarlo:2012} for a survey of contour generalizations. Cole \etal\ \cite{Cole:2008} performed a thorough study, enlisting human artists to create line drawings of known 3D objects. They found that existing methods could account for the majority of human-drawn lines, but many differences remained between hand-drawn and computer-generated lines. Gryaditskaya et al.\ \cite{OpenSketch} collect and analyze professional illustrations of objects. While these analyses yield deep insights into the nature of hand-drawn lines, the synthesis algorithms fail to match the quality of hand-drawn lines, while also requiring several parameters to be set by a user on a case-by-case basis.

Meanwhile, learned image stylization algorithms in computer vision and computer graphics, e.g., \cite{Hertzmann:2001,Gatys:2016,Zhu:2017,Johnson:2016}, have shown the potential of learning to capture artistic styles. However, these methods do not capture dependence on an underlying 3D shape, and often neglect 3D understanding. Few style transfer methods are capable of effective stylization with line drawings. Two exceptions are Im2Pencil \cite{Li:2019} and Inoue et al.\ \cite{Inoue:2019}, which separate object outlines from interiors, but do not explicitly consider 3D shape. 
StyLit stylizes 3D models \cite{Fiser16-SIG} but does not capture line drawings.

To date, there has been almost no previous work that learns artistic style for line drawings of 3D models. Lum and Ma \cite{Lum2005} proposed an SVM for learning  lines from interactive user feedback on a single shape. Kalogerakis \etal\ \cite{Kalogerakis:2012} proposed learning hatching styles from combinations of surface orientation fields and geometric features. Cole \etal\ \cite{Cole:2008} proposed linear combinations  of geometric features and decision trees to learn line drawing parameters from a small number of examples. In contrast, we learn a model that combines existing geometric and stylization algorithms, a novel large dataset, and modern neural architectures to produce a state-of-the-art line drawing algorithm.

\vspace{-1mm}
\section{Line Drawing Model}
 We first describe our architecture for computing a line drawing from a 3D shape. 
The model takes a 3D shape and camera  as input, and outputs a line drawing. The 3D shape is represented as a triangle mesh that approximates a smooth surface. The output line drawing is specified as a 2D grayscale image. Our architecture has two branches (Figure \ref{fig:architecture}): a  ``geometry branch'' that performs line drawing based on geometric features of the 3D model, and an ``image translation branch'' that learns lines through image-to-image translation.  Parameters for the geometric lines are set at run-time by a ``ranking module''.  Training for the model is described in Section \ref{sec:training}.

\subsection{Geometry branch}
\vspace{-1mm}
 The first branch of our model is based on classic geometry-based line drawing definitions, namely suggestive contours, ridges, valleys, and apparent ridges. Given a camera viewpoint and 3D shape, each of these formulations contributes to a grayscale pixel intensity map, which are combined to produce the final image $\bI$.
Their contributions depend on a set of thresholding parameters. Instead of setting these by hand, we introduce differentiable formulations to allow learning the thresholding parameters.

\begin{figure}[t]
\begin{center}
\includegraphics[width=\linewidth]{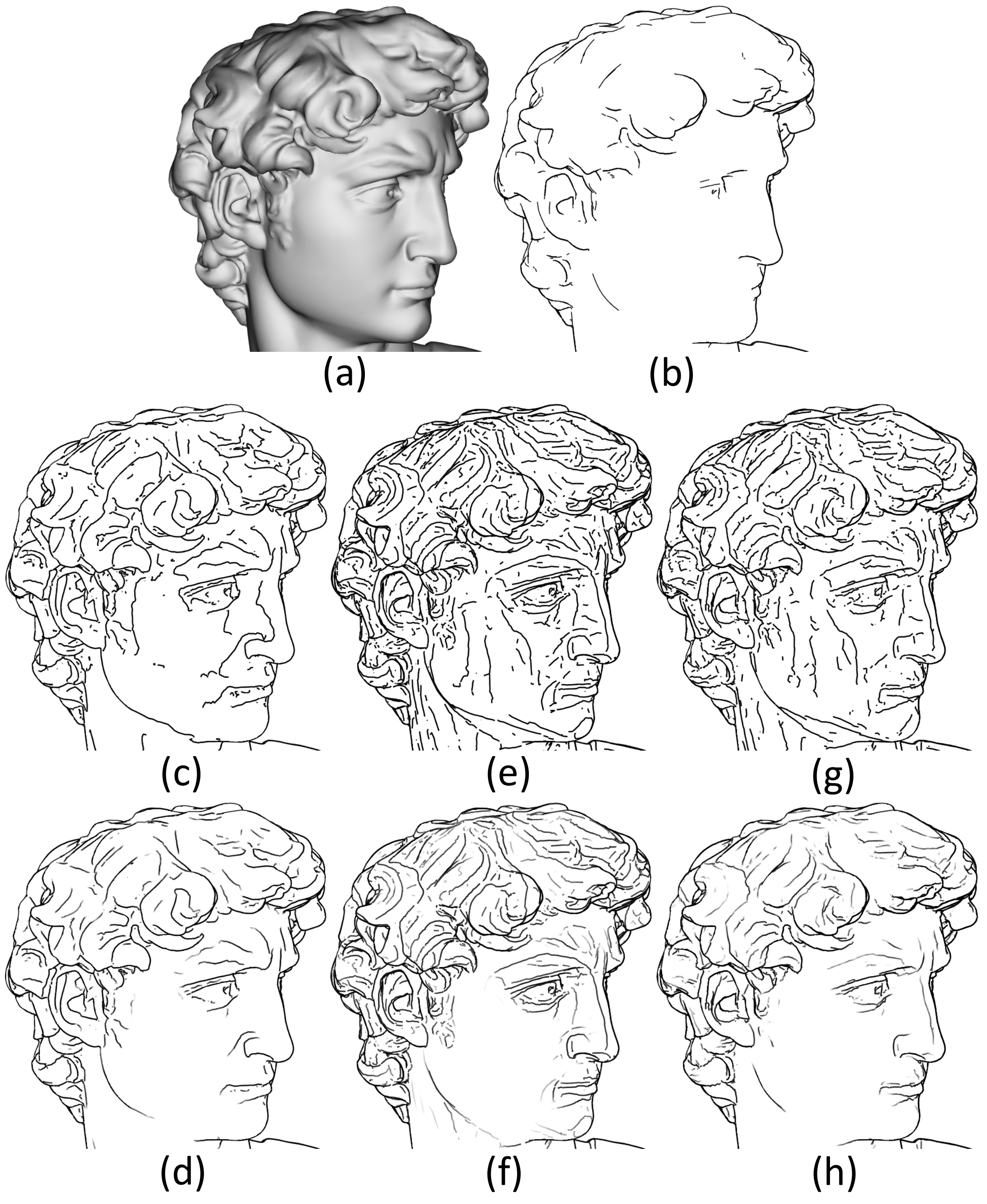}
\end{center}
\vspace{-6mm}    
   \caption{Given a 3D shape (a), we show (b) occluding contours, (c,d) unfiltered and filtered suggestive contours,  (e,f) unfiltered and filtered ridges and valleys, (g,h) unfiltered and filtered apparent ridges.} 
\vspace{-4mm}    
\label{fig:npr_example}
\end{figure}

The first set of curves produced are \textbf{Occluding Contours} \cite{BenardHertzmann}.
The model generates a binary mask $\bI_C$ with ``on'' pixels at projections of occluding contours (Figure \ref{fig:npr_example}b), computed using the interpolated contour algorithm \cite{Hertzmann:2001}.
Occluding contours are parameter-free, and do not require any learning; they are used in all of our renderings. Another parameter-free set of lines are mesh boundaries, also rendered as a binary mask  $\bI_B$. 

\begin{figure*}[t]
\begin{center}
\includegraphics[width=\textwidth]{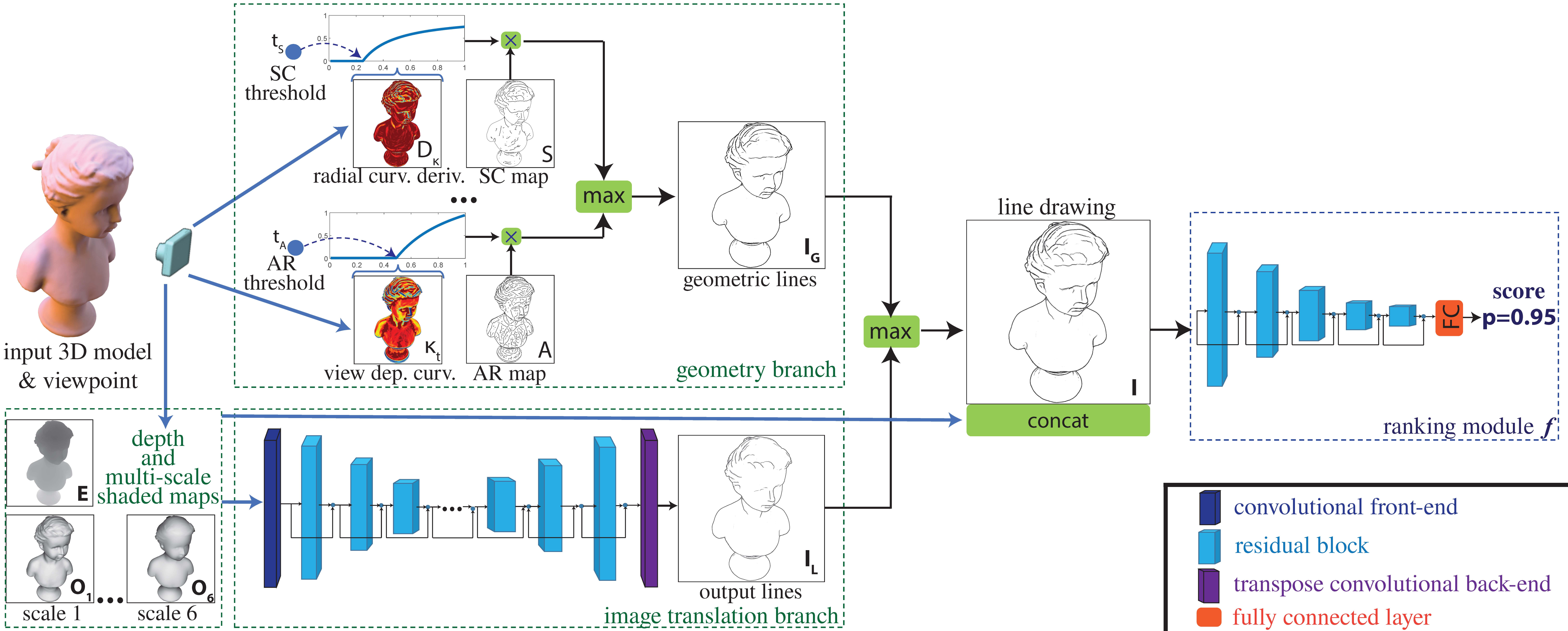}
\end{center}
\vspace{-6mm}
   \caption{Our network architecture: the input 3D model is processed by a geometry branch operating on curvature features, and an image-based branch operating on view-based representations. Their outputs are combined to create a line drawing, which is in turn evaluated by a ranking module that helps determining optimal line drawing parameters.} 
\vspace{-3mm}   
\label{fig:architecture}
\end{figure*}

\vspace{-4mm}
\paragraph{Suggestive Contours (SCs)} 
 \cite{DeCarlo:2003} represent surface points that are occluding contours in nearby views. 
 See DeCarlo \cite{DeCarlo:2003} for a detailed explanation and definitions.
Let $\kappa$ be the \textit{radial curvature}, and $D \kappa$ be the directional derivative of the radial curvature at a surface point, as defined in \cite{DeCarlo:2003}.  SCs are points where $\kappa=0$ and $D \kappa>0$. For meshes, these curves are computed by interpolation to find the zero set of $\kappa$.
As seen in Figure \ref{fig:npr_example}c, rendering all SCs is undesirable. Instead, ``weak'' SCs are filtered by only rendering SCs with $D \kappa>t_S$ for some threshold $t_S$, and tapered off below $t_S$ (Figure \ref{fig:npr_example}d) \cite{DeCarlo:2003}. In previous work, this $t_S$ parameter is manually adjusted  for each 3D model.
In order to determine this threshold automatically, we introduce a formulation that is differentiable with respect to $t_S$. For a given threshold, the method outputs a per-pixel intensity map $\bI_S$.
We build two image-space maps. First, $S(\bx)$ is a binary map that is 1 at the projections of suggestive contours, and 0 otherwise, where $\bx$ indexes pixel locations in the image. Second, $D\kappa(\bx)$ associates each pixel $\bx$ to the directional derivative of the radial curvature at the surface point visible from that pixel. Figure \ref{fig:architecture} shows these two image-space maps for an input 3D shape. 
Then, the SC image is computed for each pixel $\bx$ as:
\vspace{-2mm}
\begin{equation}
    \bI_S(\bx,t_S) = 
    S(\bx) \ \max \left(1 - \frac{t_S}{ D \kappa(\bx)},\ 0 \right)
\end{equation}
The second term filters out lines with small $D\kappa$. For $t_S=0$, all suggestive contours are displayed, while, as $t_S$ increases, they are eliminated.  The inverse function is used 
 rather than a linear dependence, e.g., $\max(D \kappa(\bx)-t_S,0)$, to produce a sharper tapering, following the implementation in \texttt{rtsc} \cite{rtsc}.
DeCarlo \etal \cite{DeCarlo:2003} also proposed filtering according to the radial direction magnitude, but we did not find that it was much different.

\vspace{-4mm}
\paragraph{Ridges and Valleys (RVs)} are viewpoint-independent surface extrema;
see \cite{ohtake2004ridge} for a detailed explanation. 
As with SCs, we introduce a formulation that is differentiable with respect to the filtering function.
We introduce a threshold for ridges ($t_R$) and one for valleys ($t_V$). The per-pixel intensity maps showing locations of ridges and valleys are generated as $R(\bx)$ and $V(\bx)$, along with maps $\kappa_1(\bx)$, $\kappa_2(\bx)$ containing the two principal curvatures of the surface point visible from each pixel, respectively.
Ridges and valleys are then filtered as:
\vspace{-2mm}    
\begin{align}
\bI_R(\bx,t_R)&=  R(\bx) \ \max \left(1.0 - \frac{ t_R } { \kappa_1 ( \bx ) },\ 0.0 \right) \\
\bI_V(\bx,t_V)&=  V(\bx) \ \max \left(1.0 - \frac{ t_V } { \kappa_2 ( \bx ) },\ 0.0 \right)
\end{align}
The interpretation of the formula is similar to SCs and yields RVs consistent with \texttt{rtsc} \cite{rtsc}. Figures \ref{fig:npr_example}e and  \ref{fig:npr_example}f  show an example of RVs before and after filtering.

\vspace{-4mm}
\paragraph{Apparent Ridges (ARs)} \cite{Judd:2007} are object ridges from a given camera position, e.g., object points that ``stick out’’ to the viewer; see \cite{Judd:2007} for a detailed description.
We define $A(\bx)$ as the map containing ARs, and filter by the view-dependent curvature $\kappa_t(\bx)$:
\vspace{-2mm}    
\begin{equation}
\bI_A(\bx,t_A)=  A(\bx) \ \max \left(1.0 - \frac{ t_A } { \kappa_t (\bx) }, 0.0 \right)
\end{equation}
Figures \ref{fig:npr_example}g and  \ref{fig:npr_example}h show ARs before and after filtering.

\vspace{-4mm}
\paragraph{Line drawing function.} Given each of these functions, we define a combined \textit{geometric line drawing function} $\bI_G$ conditioned on the set of parameters  $\bt=\{t_S, t_R, t_V, t_A\}$ (we drop the pixel id $\bx$ for clarity): 
\begin{equation}
\bI_G(\bt) =  \max \! \big( \bI_S(t_S), \bI_R(t_R), \bI_V(t_V), \bI_A(t_A), \bI_C, \bI_B \!\big)
\label{eqn:scrvapr_drawing}
\end{equation}
where the $\max$ function operates per pixel independently. 
\vspace{-4mm}
\paragraph{Preprocessing.} In a pre-processing step, we compute the curvatures required for the above lines from the input mesh.  
using \cite{Rusinkiewicz04}. We  normalize object size so that the longest dimension is equal to $1$ and the curvature quantities are divided by the their $90th$ percentile value. 

\subsection{Image translation branch}
\vspace{-1mm}
\label{sec:encoder_decoder}

An alternative approach to create line drawings from shapes is to use a neural network that directly translates shape representations to 2D images. To simplify the mapping, one can feed view-based shape representations as input to such network (i.e., depth images, shaded renderings), which also allows us to re-purpose existing  image-to-image translation networks. Our method also incorporates this generative approach. Specifically, following pix2pixHD \cite{wang2018high}, we used an image translation network module, shown in Figure~\ref{fig:architecture}. Details about the architecture are provided in the appendix. As input to the network, we use a combination of view-based representations. First, we use  a depth image $\bE$ of the shape from the given camera. Then we also compute shaded rendering images 
representing Lambertian reflectance (diffuse shading) \cite{Phong:1975:ICG}  created from the dot product of surface normals with light direction (light is at the camera). \rev{We use these shaded renderings because shading features are important predictors of where people draw lines \cite{Cole:2008}.}
To increase robustness to rendering artifacts, we also smooth the mesh normals using diffusion \cite{Jones:2003:NFM} with different smoothing parameters (more details in the appendix). As a result, we create a stack of six smoothed versions of shaded images $\bO=\{\bO_1,\bO_2,...,\bO_6\}$, which are concatenated channel-wise with the depth image  (Figure~\ref{fig:architecture}, bottom left). The resolution of all images is set to $768 \times 768$.
 We found that using these combined multiple inputs produces better results.  
 
 We also experimented with feeding rendered curvature maps, however, as we discuss in our experiments section, the results did not improve. 
Finally, since the network outputs per-pixel line probabilities, we  use the ridge detection procedure of Cole \etal \cite{Cole:2008} so that the output image $\bI_L$ contains cleaner curves.

\subsection{Neural Ranking Module}
\label{sec:ranking_module}
\label{sec:neural_ranker}
As discussed earlier, the thresholding parameters  $\bt$  play an important role in determining the existence and tapering of the geometric lines. The threshold values determine how much intensity each geometric line will contribute to the final image, if at all. 
Our approach determines the threshold parameters $\bt$ at test time, since different 3D models may be best rendered by different combinations of geometric lines. 
We employ a Neural Ranking Module (NRM) that scores the quality of a given line drawing. Then, at test time, the thresholds $\bt$ are set by optimization of the NRM score.

Specifically, the module is a function of the merged line drawing $\bI(\bt)= \max(\bI_G(\bt), \bI_L)$, the depth image of the shape from the given viewpoint $\bE$, and also the multi-scale shaded images  $\bO$ (Figure \ref{fig:architecture}). 
The module is a neural network  $f(\bI(\bt),\bE,\bO,\bphi)=p$, where $p$ is the output score, and $\bphi$ are the learned network parameters. 
At test time, we aim to maximize this function (i.e., the quality of the drawing) by modifying the parameters $\bt$:
\vspace{-1mm}
\begin{equation} 
\argmax\limits_\bt f(\bI(\bt),\bE,\bO)
\label{eqn:optimization}
\end{equation}
\rev{The maximization is done with L-BFGS using analytic gradients \mbox{$(\partial f/\partial \bI) \cdot (\partial \bI / \partial \bt)$}, computed from backpropagation  since our modules are differentiable. We also impose a non-negativity constraint on the parameters $\bt$, as specified by the geometric definitions of the lines.
To avoid local minima, we try multiple initializations of the parameter set $\bt$ through a grid search.}

The function \ref{eqn:optimization} is also used to choose whether to render mesh boundaries $\bI_B$. This is simply a binary check that passes if the function value is higher when boundaries are included in $\bI(t)$. Once the ranking network has determined the optimal parameters $\bt_{opt}$, the final drawing is output as \mbox{ $\bI(\bt) = \max(\bI_G(\bt_{opt}), \bI_L)$.} We note that the NRM  does not adjust the contribution of the image translation module at test time. We tried using a soft thresholding function on its output $\bI_L$, but it did not help. Instead, during training, the image translation module is fine-tuned with supervisory signal from the NRM.
\rev{We also experimented with directly predicting the parameters $\bt$ with a feed-forward
network, but we found that this strategy resulted in much worse performance compared to
test-time optimization of $\bt$; see the appendix for details.}

\vspace{-3mm}
\paragraph{NRM architecture.} The neural ranking module follows the ResNet-34 architecture. The input is the line drawing $\bI$, depth $\bE$, and shaded maps $\bO$  at $768 \times 768$ resolution that are concatenated channel-wise. 
To handle this input,  we added one more residual block after the original four residual blocks of ResNet-34 to downsample the feature map by a factor of 2. The newly added residual block produces a $12 \times 12 \times 1024$  map. After mean pooling, we get a $1024$-dimension feature vector. We remove the softmax layer of ResNet-34 and use a fully connected layer to output the ``plausibility'' value. Details about the architecture  can be found in the appendix.

\section{Dataset}
\label{sec:training}
\vspace{-1mm}
To train the the neural ranking  and  image translation modules, we need a dataset of line drawings. Although there are a few large-scale human line drawing datasets available online \cite{quickdraw,Sangkloy:2016:SDL},  the drawings are not associated  to reference 3D shapes and include considerable distortions.
An alternative scenario is to ask artists to provide us with line drawings depicting training 3D
shapes. However, gathering a large number of human line drawings for training is labor-intensive and time-consuming. Cole \etal's dataset \cite{Cole:2008} was gathered this way, and is too small on its own to train a deep model.
In contrast, for each training shape, we generated multiple synthetic line drawings using \texttt{rtsc} \cite{rtsc} through several combinations of different lines and thresholds, and asked human subjects to select the best drawing in a relative comparison setting. Since selecting the best drawing can be subjective, we gathered votes from multiple human subjects, and used only training drawings for which there was consensus. 
Below we describe our dataset, then we describe the losses  to train our modules.

\vspace{-3mm}
\paragraph{Shape dataset.} The first step to create our dataset was to select training 3D shapes  from which  reference line drawings will be generated.
We used three collections: ShapeNet \cite{chang2015shapenet},  Models Resource \cite{modelsresource}, and Thingi10K \cite{zhou2016thingi10k}.  These shape collections contain a large variety of human-made and organic objects. In the case of ShapeNet, we sampled a random subset of up to $200$ shapes from each category, to avoid category imbalance. 
All models have oriented ground planes specified.
We  removed duplicate shapes and avoided sampling low-resolution shapes with fewer than  2K faces. 
We also processed the meshes by
 correctly  orienting  polygons  (front-facing with respect to   external viewpoints), and repairing connectivity (connect geometrically adjacent but topologically disconnected polygons, weld coincident vertices).
The  number of shapes in all collections is 23,477.

\vspace{-3mm}
\paragraph{Generating candidate line drawings.}
 We select two random camera positions for each 3D model under the constraint that they are elevated $30$ degrees from the ground plane, are aligned with the upright axis, and they point towards the centroid of the mesh (i.e., only the azimuth of each camera position is randomized).
Then for each of the two camera positions, we generated 256 synthetic line drawings using all possible combinations of suggestive contours, apparent ridges, ridges with valleys
under 4 thresholds (e.g. suggestive contours have 4 different thresholds [0.001, 0.01, 0.1, off]), including combinations with and without mesh creases and  borders ($4 \times 4 \times 4 \times 2 \times 2 = 256$ combinations). We also generated additional synthetic line drawings using Canny edges and edge-preserving filtering \cite{gastal2011domain} on rendered shaded images of shapes, each with  4 different edge detection thresholds resulting in $8$ more drawings. In total, this process resulted in 264 line drawings for each shape and viewpoint.  These line drawings can be similar to each other, so we selected the 8 most distinct ones by applying k-mediods ($k=8$) and using Chamfer distance between drawn lines as metric to recover the clusters.

\begin{figure}[t]
\begin{center}
\includegraphics[width=\linewidth]{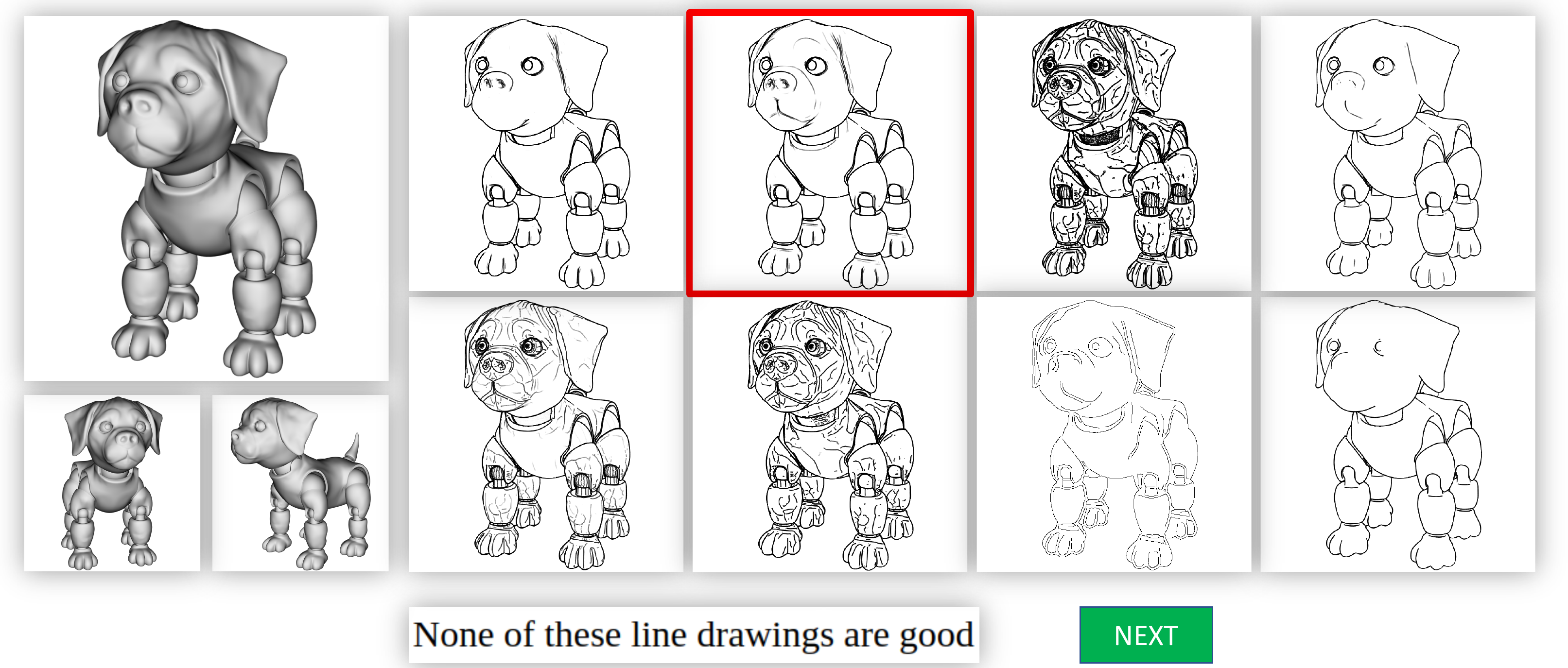}
\end{center}
\vspace{-6mm} 
   \caption{A snapshot from our MTurk questionnaires used for gathering training line drawing comparisons. The most voted answer is highlighted as red.}
\vspace{-4mm}    
\label{fig:interface}
\end{figure}

\vspace{-3mm}
\paragraph*{Questionnaires.} We then created Amazon MTurk questionnaires, where, on each page,  we showed  $8$ candidate line drawings of a 3D shape, along with rendered images of it from different viewpoints \cite{wilber2014cost} (Figure \ref{fig:interface}). Each page asked human participants to select the line drawing that best conveyed the rendered shape, and was most likely to be selected by other people as well. 
We also provided the option not to select any drawing (i.e., if none depicted the shape reliably).  We employed sentinels (drawings of irrelevant shapes) to filter unreliable participants. We had total $3,739$ reliable MTurk participants in our study (see appendix for details). For each shape and viewpoint, we gathered votes from 3 different participants. We accepted a drawing for training if it was chosen by at least two users. As a result, we gathered $21,609$ training line drawings voted as ``best'' per shape and viewpoint. A random subset ($10\%$ of the original dataset) was kept for hold-out validation.

\section{Training}
\label{sec:training}

The goal of our training procedure is to learn the parameters $\bphi$ of the neural ranking module, and the parameters $\btheta$ of the image translation branch from our training dataset, so that high-quality drawings can be generated for 3D\ models.

\vspace{-3mm}
\paragraph{NRM training.} To train the Neural Ranking Module $f$, we use a ranking loss   based on above crowdsourced comparisons. Given a drawing $\bI_{\mathit{best}}^{s,c}$ selected as ``best'' for shape $s$ and viewpoint $c$, we generate $7$ pairwise comparisons consisting  of that drawing and every other drawing $\bI_{\mathit{other},j}^{s,c}$ that participated in the questionnaire. We use the hinge ranking loss to train the module \cite{Herbrich99,Dehghani:2017:NRM}:
\begin{align}
    L_R \!=\!\sum\limits_{s,c,j} \max(m &- f(\bI_{\mathit{best}}^{s,c},\bE,\bO,\bphi) \nonumber \\ 
   & +f(\bI_{\mathit{other},j}^{s,c},\bE,\bO,\bphi) , 0)
\end{align}
where $m$ is the margin set to $1.0$.

\vspace{-3mm}
\paragraph*{Image translation module training.} Based on the line drawings  selected as ``best'' per reference shape and viewpoint, we use  cross-entropy to train the image translation module. Specifically, we treat the intensity values $\bI_{best}$ as target probabilities for drawing, and measure the cross-entropy  of the predicted output $\bI_L$ and $\bI_{best}$:
\begin{equation}
L_{ce} \!=\! -\!\sum\limits_x(\bI_{best}(x) \log\bI_{L}(x) +(1 \!-\! \bI_{best}(x)) \log(1 \!-\!\bI_{L}(x))) \nonumber
\end{equation}
\rev{We then further end-to-end fine-tune the image translation module along with the NRM\ module based on the ranking loss.}
We also experimented with adding the GAN-based losses of pix2pix \cite{pix2pix2016} and pix2pixHD \cite{wang2018high}, yet, their main effect was only a slight sharpening of existing lines, without adding or removing any new ones.

\vspace{-4mm}
\paragraph*{Implementation.} For the ranking module, we used the Adam optimizer \cite{kingma2014adam} with learning rate $2\cdot10^{-5}$ and batch size $32$. For the image translation module, we used Adam with learning rate set to $2\cdot10^{-4}$ and batch size $2$.  The PyTorch implementation and our dataset are available at:\\
\small{\url{http://github.com/DifanLiu/NeuralContours}}

\section{Results}
\label{sec:results}

We evaluate our method and alternatives quantitatively and qualitatively. To perform our evaluation, we compare  synthesized line drawings with ones drawn by humans for reference shapes.
Below, we
describe our test datasets, evaluation measures, and comparisons with  alternatives. 

\vspace{-3mm} 
\paragraph{Test Datasets.} Cole \etal~\cite{Cole:2008}  conducted a study in which artists made line drawings intended to convey given 3D shapes. The dataset contains $170$ precise human line drawings of $12$ 3D models under different viewpoints and lighting conditions for each model. Their resolution is $1024 \times 768$ pixels. Since the number  of 3D\ test models is small, we followed the same setup as Cole \etal~to gather $88$ more human line drawings from $3$ artists for $44$ 3D models under two viewpoints (same resolution), including printing renderings, scanning their line drawings, and aligning them.  Our new test dataset includes $13$ 3D animal models, $10$ human body parts, 
$11$ furniture,  $10$ vehicles and mechanical parts. All 3D models (including the ones from Cole \etal's dataset) are disjoint from the training and validation sets.

\vspace{-3mm} 
\paragraph*{Evaluation measures.} 
We use precision and recall measures for comparing synthetic drawings to human-made drawings, computed in the manner proposed by Cole \etal \cite{Cole:2008}. Each drawing is first binarized through thinning and thresholding.  Precision is defined as the fraction of drawn pixels in a synthetic drawing that are near any drawn pixel of the human drawing of the same shape under the same viewpoint. Recall is defined as the fraction of pixels in the human drawing that are near any line of the synthetic drawing. Two pixels are ``near'' if they are within 1mm in the coordinates of the physical page the drawing was made on; this distance was based on measurements of agreement between human drawings in Cole \etal's dataset.
We aggregate precision and recall into F1-score.

We also report Intersection over Union (IoU) to measure overlap between synthetic and human drawings, based on the same definition of nearness. Lastly, we report the symmetric Chamfer distance, which  measures the average distance between lines of synthetic and human drawings.
Since both human and synthetic    drawings include aligned silhouettes, all measures will appear artificially improved because of them. To avoid this biasing, we  remove silhouettes from all human and synthetic drawings and measure performance based on the rest of the lines only.

\begin{figure*}
\begin{center}
\includegraphics[width=\linewidth]{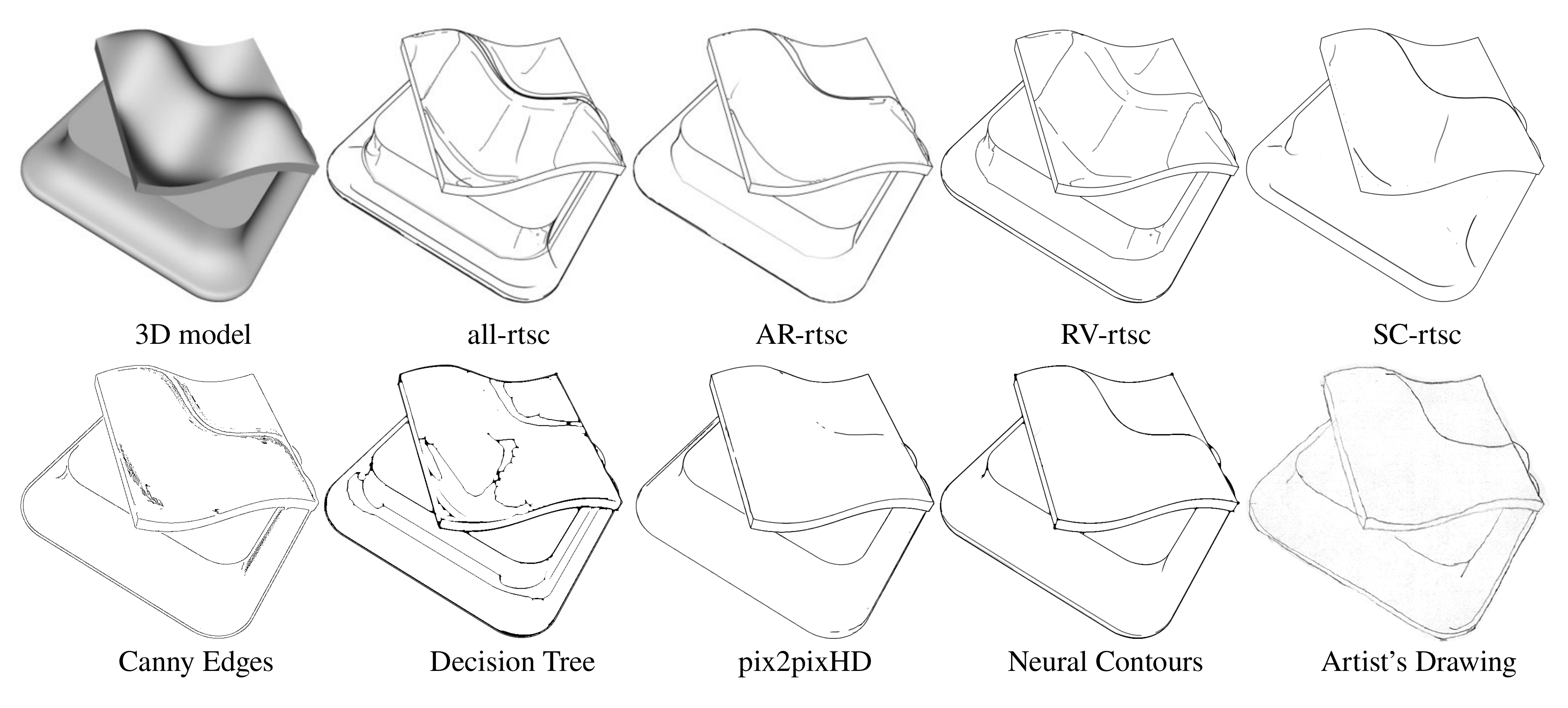}
\end{center}
\vspace{-8mm} 
   \caption{Comparisons with other methods. Neural Contours are more consistent with underlying shape features.}
\vspace{-3mm}    
\label{fig:comparisons}
\end{figure*}

\vspace{-3mm} 
\paragraph*{Comparisons.} We compare our method, Neural Contours (NCs), with several alternatives. 
(1) \emph{Occluding Contours}. 
(2-4) \emph{SC-rtsc, RV-rtsc, AR-rtsc} using  \texttt{rtsc} \cite{rtsc} with the default thresholding parameters; occluding contours are also included in all renderings.
(5) \emph{all-rtsc} renders SCs, RVs, and ARs all together with \texttt{rtsc} \cite{rtsc}, using the default parameters.
We note that we also tried to tune these parameters using an exhaustive grid search to minimize average Chamfer distance in the training set, but this resulted in worse performance (see appendix). 
(6) \emph{decision tree:} The method of Cole \etal \cite{Cole:2008}, namely, a decision tree (M5P from Weka \cite{Hall:2009:WDM})
operating on rendered curvature and gradient maps, trained on our dataset.
(7) \emph{Canny} edges extracted from the shaded shape rendering, as suggested by \cite{Cole:2008}. The edge detection parameters are selected using grid search to minimize average Chamfer distance in our training dataset. 
(8) \emph{pix2pixHD} image translation \cite{wang2018high},  trained to output line drawings from an input depth image and shaded renderings of a shape
(same as $\bE,\bO$ in our method). Training was done on the same dataset (``best'' drawings) as ours using the GAN and feature matching loss \cite{wang2018high}. The original architecture  outputs a $4096 \times 2048$ image through three local enhancer networks. In our case, since the input and output have the same resolution ($1024 \times 768$), we use only the global generator of pix2pixHD. 

\vspace{-3mm} 
\paragraph*{Ablation study.} We also compare with training the following reduced variants of our method.
\emph{NC-geometry} uses our geometry-based  branch and our neural ranker module.
\emph{NC-image} uses the image translation module alone trained with the same losses as ours, and multi-scale shaded images as input. 
\emph{NC-image-noms}  uses the image translation module alone trained
with the same losses as ours, and using a single shaded and depth image as input (no multi-scale shaded images). 
\emph{NC-curv} is an alternative image translation module that  uses curvature maps rendered in image-space concatenated with the multi-scale shaded images and depth. 

\begin{table}
\begin{center}
\begin{tabular}{|c|c|c|c||c|c|}
\hline
Method & IoU & CD & F1 & P & R  \\
\hline\hline
\emph{contours} & 31.5 & 31.70 & 34.8 & 83.0 & 22.0  \\
\emph{AR-rtsc} & 53.4 & 12.56 & 54.1 & 52.5 & 55.7  \\
\emph{RV-rtsc} & 49.8 & 12.96 & 52.3 & 44.5 & 63.5  \\
\emph{SC-rtsc} & 40.5 & 13.96 & 44.0 & 43.9 & 44.1  \\
\emph{all-rtsc} & 48.2 & 12.63 & 52.5 & 40.4 & 75.1  \\
\emph{Decision Tree} & 46.9 & 12.17 & 49.9 & 38.6 & 70.4
\\
\emph{Canny Edges} & 51.9 & 12.59 & 52.9 & 50.4 & 55.8  \\
\emph{pix2pixHD} & 45.0 & 15.73 & 48.7 & 69.6 & 37.5  \\
\cline{1-6}
\emph{NCs}& \textbf{57.9} & \textbf{10.72} & \textbf{60.6} & 60.8 & 60.5   \\

\hline
\end{tabular}
\end{center}
\vskip -6mm
\caption{Comparisons with competing methods using all drawings from Cole \etal's dataset. IoU, F1, P, R are reported in percentages, CD is pixel distance.}
\vskip -3mm
\label{table:princeton_dataset_all_drawings}
\end{table}

\begin{table}
\begin{center}
\begin{tabular}{|c|c|c|c||c|c|}
\hline
Method & IoU & CD & F1 & P & R \\
\hline\hline
\emph{contours} & 43.5 & 24.63 & 49.6 & 90.2 & 34.3  \\
\emph{AR-rtsc} & 59.9 & 10.64 & 63.3 & 62.6 & 64.0  \\
\emph{RV-rtsc} & 52.6 & 10.76 & 55.3 & 47.0 & 67.3  \\
\emph{SC-rtsc} & 46.6 & 12.27 & 50.6 & 52.1 & 49.1  \\
\emph{all-rtsc} & 51.8 & 10.74 & 56.5 & 43.8 & 79.6  \\
\emph{Decision Tree} & 49.7 & 11.12 & 53.0 & 41.1 & 74.6  \\
\emph{Canny Edges} & 58.0 & 11.16 & 61.3 & 56.7 & 66.7  \\
\emph{pix2pixHD} & 50.5 & 13.35 & 54.2 & 75.1 & 42.4  \\
\cline{1-6}
\emph{NCs}& \textbf{65.2} & \textbf{8.71} & \textbf{67.6} & 66.3 & 69.0   \\

\hline
\end{tabular}
\end{center}
\vskip -6mm
\caption{
Comparisons with other methods using the most ``consistent'' human drawings  from Cole \etal's dataset.}
\vskip -3mm
\label{table:princeton_dataset_best_drawings}
\end{table}

\begin{table}
\begin{center}
\begin{tabular}{|c|c|c|c||c|c|}
\hline
Method & IoU & CD & F1 & P & R \\
\hline\hline
\emph{contours} & 49.0 & 19.11 & 54.9 & 92.2 & 39.1  \\
\emph{AR-rtsc} & 66.8 & 9.19 & 69.9 & 69.2 & 70.7  \\
\emph{RV-rtsc} & 64.8 & 9.36 & 66.2 & 62.8 & 70.1  \\
\emph{SC-rtsc} & 65.0 & 9.88 & 63.3 & 61.5 & 65.2  \\
\emph{all-rtsc} & 64.4 & 9.70 & 68.6 & 58.6 & 82.7  \\
\emph{Decision Tree} & 62.1 & 8.93 & 61.1 & 50.9 & 76.6  \\
\emph{Canny Edges} & 65.6 & 8.57 & 64.6 & 59.8 & 70.2  \\
\emph{pix2pixHD} & 66.0 & 9.62 & 68.2 & 76.9 & 61.2  \\
\cline{1-6}
\emph{NCs}& \textbf{72.4} & \textbf{7.25} & \textbf{74.6} & 74.5 & 74.8   \\

\hline
\end{tabular}
\end{center}
\vskip -6mm
\caption{Comparisons in our new test dataset.}
\vskip -1mm
\label{table:new_dataset}
\end{table}

\begin{table}
\begin{center}
\begin{tabular}{|c|c|c|c||c|c|}
\hline
Method & IoU & CD & F1 & P & R \\
\hline\hline
\emph{\small{NC-geometry}} & 60.3 & 10.34 & 64.5 & 76.9 & 55.6  \\
\emph{\small{NC-image}} & 60.0 & 9.97 & 62.9 & 65.0 & 61.0  \\
\emph{\small{NC-image-noms}} & 58.4 & 10.85 & 60.7 & 59.1 & 62.3  \\
\emph{\small{NC-image-curv}} & 56.1 & 10.72 & 60.0 & 61.0 & 59.0  \\
\cline{1-6}
\emph{NCs}& \textbf{62.8} & \textbf{9.54} & \textbf{65.4} & 65.5 & 65.4  \\

\hline
\end{tabular}
\end{center}
\vskip -6mm
\caption{Ablation study.}
\vskip -3mm
\label{table:ablation}
\end{table}

\vspace{-3mm} 
\paragraph*{Results.}
  Tables \ref{table:princeton_dataset_all_drawings} reports the evaluation measures for Cole \etal's dataset for competing methods. Specifically, the synthetic drawings are compared with each human line drawing per shape and viewpoint, and the measures are averaged. Since there are artists that draw more consistently than others, we also include Table \ref{table:princeton_dataset_best_drawings} as an alternative comparison. This table reports
the evaluation measures in Cole \etal's dataset when synthetic drawings are compared only with the most ``consistent'' human line drawing  per shape and viewpoint, defined as the drawing that has the least Chamfer distance
 to the rest of the human drawings for that shape and viewpoint. We believe this comparison is  more reliable than using all drawings, because in this manner, we disregard any ``outlier'' drawings, and  also get a measure of how well methods  match the most consistent, or agreeable, human line drawing. Table \ref{table:new_dataset} also reports the evaluation measures  for our new dataset. 
 Based on the results, Neural Contours outperforms all competing methods in terms of IoU, Chamfer Distance, and F1, especially when we compare with the most consistent human drawings.  
 
 Table \ref{table:ablation} reports comparisons with reduced variants of our method for the purpose of our ablation study. 
Our two-branch architecture offers the best performance compared to using  the  individual branches alone. 
 
\begin{figure}[t]
\begin{center}
\includegraphics[width=\linewidth]{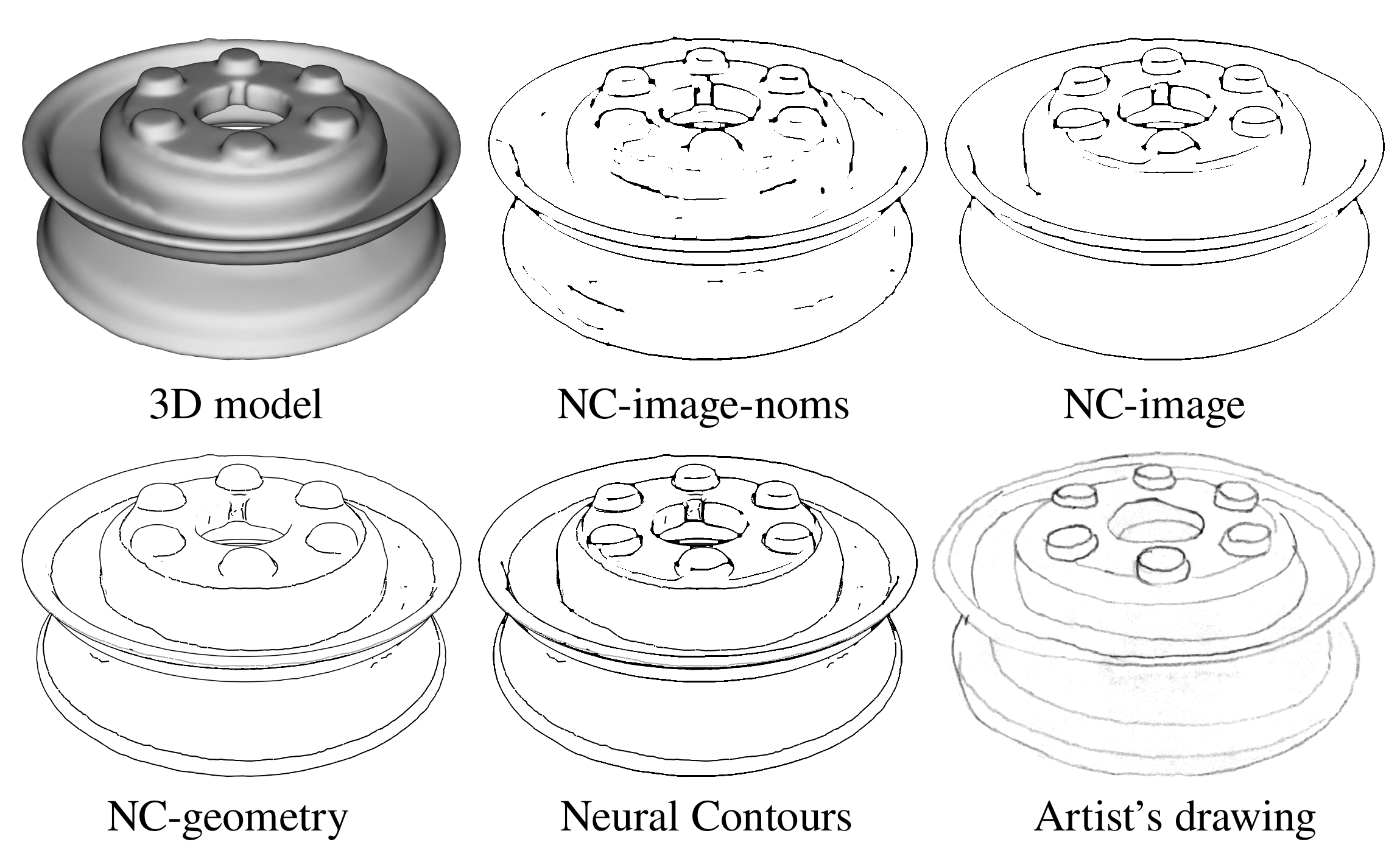}
\end{center}
\vspace{-6mm} 
   \caption{Comparisons with reduced NCs variants.}
\vspace{-2mm}    
\label{fig:ablation}
\end{figure}

 Figure \ref{fig:comparisons} shows characteristic comparisons with competing methods, and  Figure \ref{fig:ablation}  shows comparisons with reduced variants of our method. We also include representative human drawings. Both figures indicate that our full method produces lines that convey shape features more similarly to what an artist would do.  Figure \ref{fig:teaser} also shows comparisons with other alternatives. Our method tends to produce more accurate lines that are more aligned with underlying shape features, and with less artifacts.
 
 \vspace{-1mm} 
  \paragraph{Are the two branches learning the same lines?} 
  To check this hypothesis, we measure the IoU between the line drawings created from the geometry branch alone and the ones created from the image translation branch. The  average IoU is $69.4 \%$.  This indicates  that the two branches outputs have a partial overlap, but still they have substantial differences.
\rev{As shown in Figure \ref{fig:ablation}, the geometry branch makes explicit use of surface information in 3D, such as surface curvature, to identify important curves, which appear subtle or vanish in 2D rendered projections. In contrast, the image branch identifies curves that depend on view-based shading information that is not readily available in the 3D geometry (see appendix for other examples).}

\vspace{-1mm}  
\paragraph{Which geometric lines are used more in our method?} The average percentage of SCs, RVs, ARs selected by our geometry-based stylization branch are $32.2\%$, $16.5\%$, $51.3\%$ respectively. It seems that ARs are used more dominantly, while RVs are the least frequent lines.

\begin{figure}[t]
\begin{center}
\includegraphics[width=\linewidth]{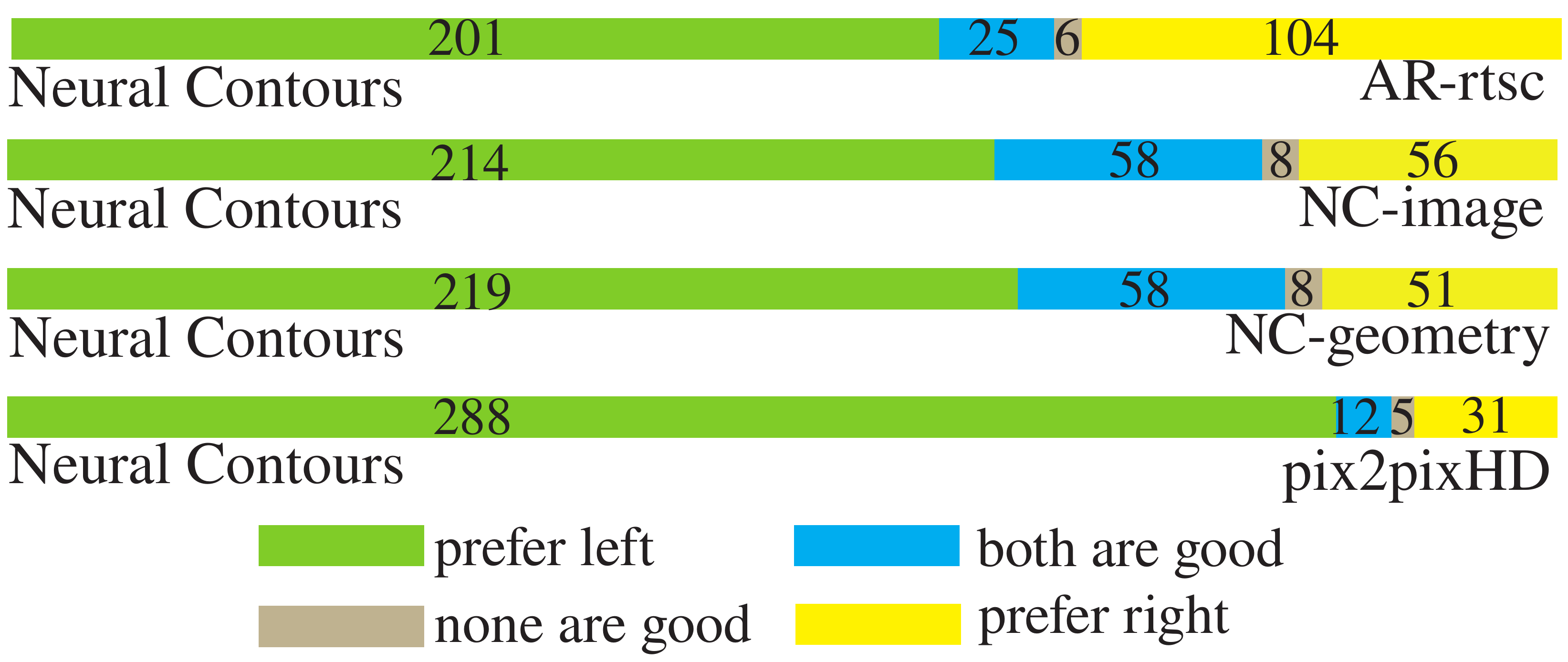}
\end{center}
\vspace{-5mm} 
   \caption{User study voting results.}
\vspace{-4mm}    
\label{fig:user_study}
\end{figure}

\vspace{-1mm} 
\paragraph{User study.} We also conducted an Amazon MTurk study as additional perceptual evaluation. Each questionnaire page showed participants shaded renderings of a shape, along with a randomly ordered pair of synthetic drawings: one synthetic drawing from our method, and another from a different one. We asked participants which drawing best conveyed the shown 3D model. Participants could pick either drawing,  specify ``none'', or ''both'' drawings conveyed the shape equally well. We asked questions twice in a random order to verify participants' reliability. We had $187$ reliable participants (see appendix for  details). 
Figure \ref{fig:user_study} summarizes the number of votes for the above options. Our method received twice the number of votes compared to the best alternative (ARs) found in this study.

\section{Conclusion}

\rev{We presented a method that learns to draw lines for 3D models based on a combination of a differentiable geometric module and an image translation network. Surprisingly, since the study by Cole \etal \cite{Cole:2008}, there has been  little progress on improving line drawings for 3D\ models.
Our experiments demonstrate that our method significantly improves over existing geometric and neural image translation methods. There are still  avenues for further improvements. Mesh artifacts (e.g., highly irregular tesselation) affect curvature estimation and shading, and in turn the outputs of both  branches. Learning to repair such artifacts to increase robustness would be fruitful.  Predicting drawing parameters in real-time is an open problem. Rendering the lines with learned pressure, texture, or thickness could make them match human drawings even more.
Finally, our method  does not handle point clouds, which would either require  a mesh reconstruction step or learning to extract lines directly from unstructured point sets.}
\vspace{-1mm}
\paragraph{Acknowledgements.} 
\rev{This  research  is   partially funded  by  NSF (CHS-1617333) and Adobe. Our  experiments  were  performed  in  the UMass GPU cluster obtained under the Collaborative Fund managed by the Massachusetts Technology Collaborative. We thank Keenan Crane for the Ogre, Snakeboard and Nefertiti 3D models.
}

{\small
\bibliographystyle{ieee_fullname}
\bibliography{egbib}
}

\renewcommand\appendixpagename{Appendix}
\appendix
\appendixpage

\section*{1. Additional Results}
\begin{figure*}
\begin{center}
\includegraphics[width=\textwidth]{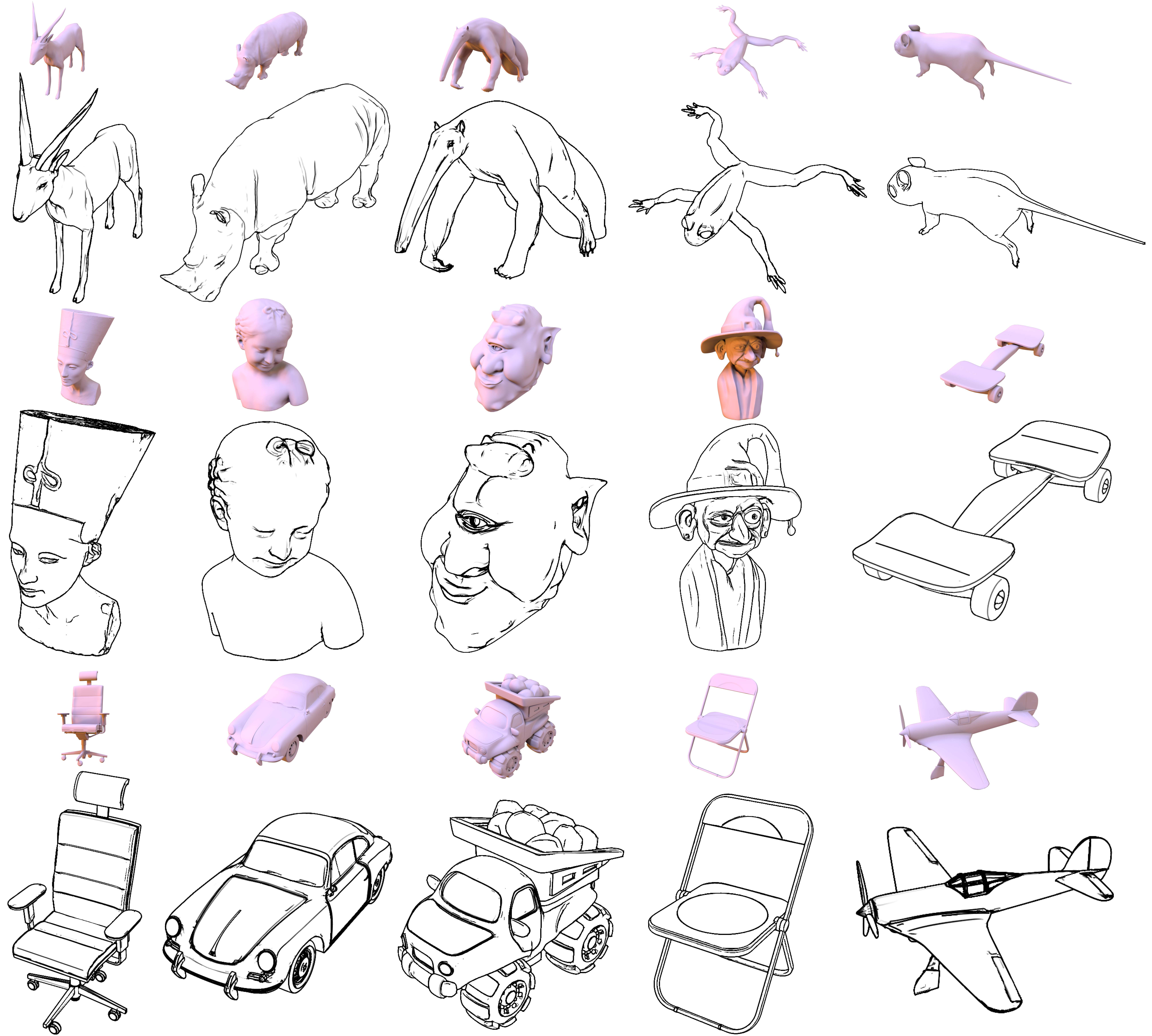}
\end{center}
\vspace{-4mm} 
   \caption{Results of our ``Neural Contours'' method on various test 3D models.}
\vspace{-2mm}    
\label{fig:supp_gallery}
\end{figure*}

Figure \ref{fig:supp_gallery} shows a gallery of our results for various 3D models from our test set (please zoom-in to see more details). We also refer the reader to more results  available on our project page:\\
\small{\url{http://github.com/DifanLiu/NeuralContours}}

\section*{2. Image Translation Branch Implementation}

We provide here more implementation details for our image translation branch (see also Section 3.2 of our main text). We also refer readers to our publicly available  implementation on our project page.

\paragraph{Multi-scale shaded maps.} 
We smooth mesh normals by diffusing the vertex normal field in one-ring neighborhoods of the mesh through a Gaussian distribution. Each vertex normal is expressed as a weighted average of neighboring vertex normals. The weights are set according to Gaussian functions on vertex distances. The standard deviation $\sigma$ of the Gaussians control the degree of influence of neighboring vertex normals: when $\sigma$ is large, the effect of smoothing is larger. The map $\bO_1$ is generated based on the initial normal field, while $\bO_2, \bO_3, \bO_4, \bO_5, \bO_6$ are created using smoothing based on $\sigma=\{1.0, 2.0, 3.0, 4.0, 5.0\}$ respectively.

\paragraph{Architecture details.} Our image translation branch uses the architecture shown in Table \ref{table:arch_ITB}. All convolutional layers are followed by batch normalization and a ReLU nonlinearity except the last convolutional layer. The last convolutional layer is followed by a sigmoid activation function. The branch contains $9$ identical residual blocks, where each residual block contains two $3 \times 3$ convolutional layers with the same number of filters for both layers. 
\begin{table}[h!]
\begin{center}
\begin{tabular}{|c|c|}
\hline
\textbf{Layer} & \textbf{Activation size}  \\
\hline\hline
Input & 768 $\times$ 768 $\times$ 7  \\
Conv(7x7, 7$\rightarrow$64, stride=1)   & 768 $\times$ 768 $\times$ 64  \\
Conv(3x3, 64$\rightarrow$128, stride=2)   & 384 $\times$ 384 $\times$ 128  \\
Conv(3x3, 128$\rightarrow$256, stride=2)   & 192 $\times$ 192 $\times$ 256  \\
Conv(3x3, 256$\rightarrow$512, stride=2)   & 96 $\times$ 96 $\times$ 512  \\
Conv(3x3, 512$\rightarrow$1024, stride=2)   & 48 $\times$ 48 $\times$ 1024  \\
9 Residual blocks   & 48 $\times$ 48 $\times$ 1024  \\
Conv(3x3, 1024$\rightarrow$512, stride=1/2)   & 96 $\times$ 96 $\times$ 512  \\
Conv(3x3, 512$\rightarrow$256, stride=1/2)   & 192 $\times$ 192 $\times$ 256  \\
Conv(3x3, 256$\rightarrow$128, stride=1/2)   & 384 $\times$ 384 $\times$ 128  \\
Conv(3x3, 128$\rightarrow$64, stride=1/2)   & 768 $\times$ 768 $\times$ 64  \\
Conv(7x7, 64$\rightarrow$1, stride=1)   & 768 $\times$ 768 $\times$ 1  \\
\hline
\end{tabular}
\end{center}
\vskip -6mm
\caption{Architecture of the Image Translation Branch.}
\vskip -3mm
\label{table:arch_ITB}
\end{table}

\section*{3. Neural Ranking Module Implementation}
We provide here  implementation details for our Neural Ranking Module (see also Section 3.3 of our main text).

\paragraph{Architecture details.} Our Neural Ranking Module uses the architecture shown in Table \ref{table:arch_NRM}. It follows
the ResNet-34 architecture. We add one more residual block with $1024$ filters after the original four residual blocks. After average pooling, we get a $1024$-dimensional feature vector. We remove the softmax layer of ResNet-34 and use a fully connected layer to output the “plausibility” value.

\begin{table}[h!]
\begin{center}
\begin{tabular}{|@{}c@{}|c|}
\hline
\textbf{Layer} & \textbf{Activation size}  \\
\hline\hline
Input & 768 $\times$ 768 $\times$ 3  \\
Conv(7x7, 8$\rightarrow$64, stride=2)   & 384 $\times$ 384 $\times$ 64  \\
Max-pool(3x3, stride=2)   & 192 $\times$ 192 $\times$ 64  \\
ResBlock(64$\rightarrow$64, stride=1, blocks=3)   & 192 $\times$ 192 $\times$ 64  \\
ResBlock(64$\rightarrow$128, stride=2, blocks=4)   & 96 $\times$ 96 $\times$ 128  \\
ResBlock(128$\rightarrow$256, stride=2, blocks=6)   & 48 $\times$ 48 $\times$ 256  \\
ResBlock(256$\rightarrow$512, stride=2, blocks=3)   & 24 $\times$ 24 $\times$ 512  \\
ResBlock(512$\rightarrow$1024, stride=2, blocks=3)   & 12 $\times$ 12 $\times$ 1024  \\
Average-pool(12x12)   & 1024  \\
FC(1024$\rightarrow$1)   & 1  \\
\hline
\end{tabular}
\end{center}
\vskip -6mm
\caption{Architecture of the Neural Ranking Module.}
\vskip -3mm
\label{table:arch_NRM}
\end{table}

\section*{4. Additional Experiments}

\paragraph{Parameter set $\bt$ regression.}
We experimented with directly predicting the parameter set $\bt$ with a network, but this did not produce good results. The network includes a mesh encoder which is a graph neural network based on NeuroSkinning and an image encoder based on ResNet-34. The mesh encoder takes a triangular mesh as input and outputs a $1024$-dimensional feature vector. The image encoder takes ($\bE,\bO$) as input and outputs a $1024$-dimensional feature vector. These two feature vectors are concatenated and processed by a $3-$layer MLP which outputs the parameter set $\bt$. We used cross-entropy loss between $\bI_G(\bt)$ and $\bI_{best}$ to train the network. We note that combining the mesh and image encoder worked the best. We name this variant \emph{Geometry-Regressor}. Table \ref{table:supp_exp} reports the resulting performance compared to our method. The results of this approach are significantly worse.

\paragraph{Parameter set $\bt$ exhaustive search.}
We also tried to tune parameters of ARs, RVs, SCs using an exhaustive grid search to minimize average Chamfer distance in the training set. The grid was based on regular sampling $100$ values of the parameters in the interval $[0,1]$. This exhaustive search did not produce good results. Table \ref{table:supp_exp} reports the performance of these variants \emph{AR-grid}, \emph{RV-grid}, \emph{SC-grid}, \emph{all-grid}.

\begin{figure}[h!]
\begin{center}
\includegraphics[width=1.0\linewidth]{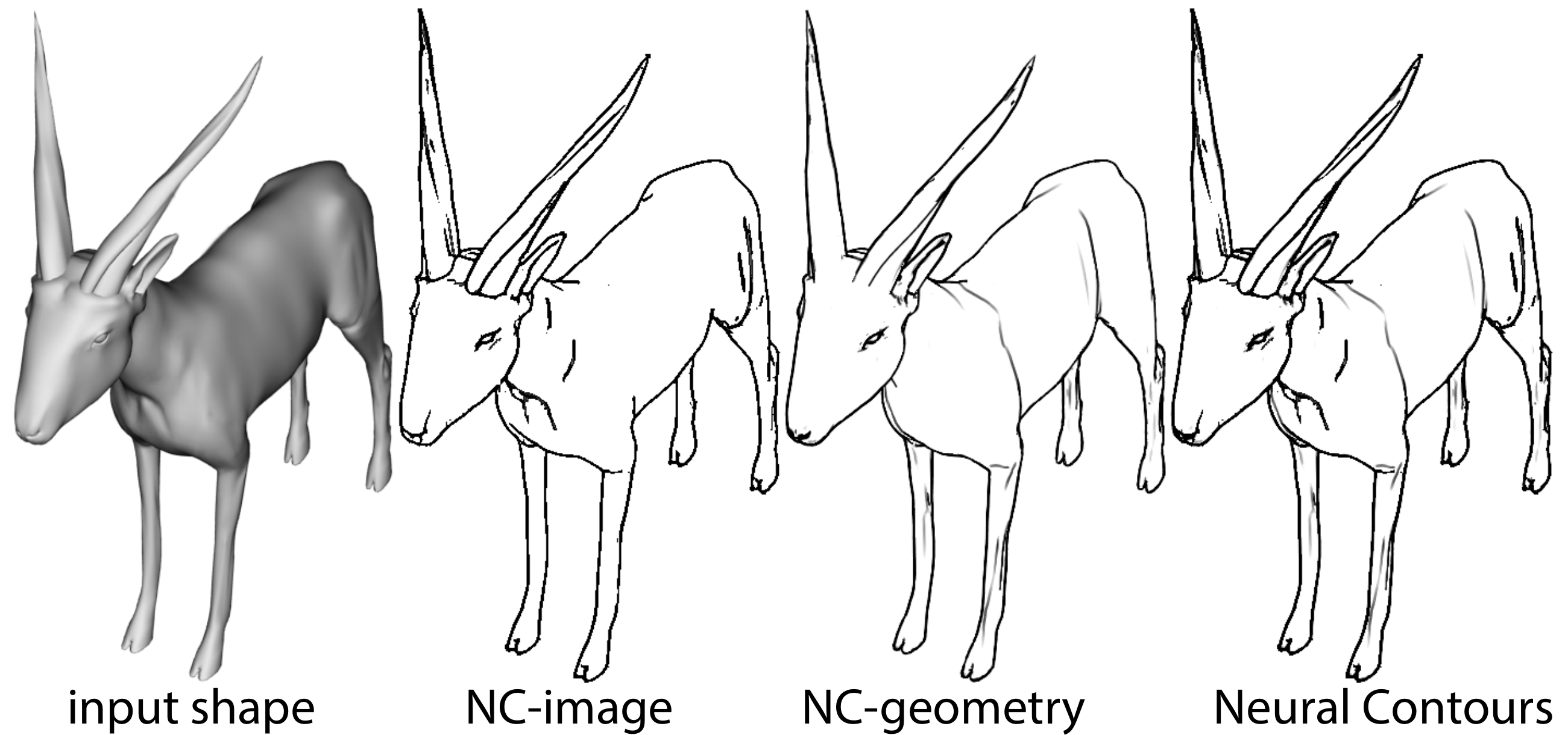}
\end{center}
\vspace{-5mm}
\caption{Additional comparison of our two branch outputs (image translation branch output ``NC-Image'' vs geometry branch output ``NC-Geometry'' vs Neural Contours).}
\label{fig:R1}
\end{figure}

\paragraph{Image
translation vs geometric branch output example}
 Figure \ref{fig:R1}
shows an additional example of comparison between the geometry branch and the image translation branch outputs; compare the areas around the antlers, and the shoulder to see the contributions of each branch. 

 As also shown in Figure 6 of our main paper, the geometry model
makes explicit use of surface information in 3D, such as surface curvature, to identify important curves, which appear subtle or vanish in 2D rendered projections. In contrast, the image model identifies curves that depend on view-based shading information that is not readily available in the 3D geometry.
\begin{table}[h!]
\begin{center}
\begin{tabular}{|c|c|c|c||c|c|}
\hline
Method & IoU & CD & F1 & P & R \\
\hline\hline
\emph{AR-grid} & 56.6 & 11.21 & 59.1 & 54.2 & 64.9  \\
\emph{RV-grid} & 56.0 & 11.73 & 58.3 & 53.6 & 63.9  \\
\emph{SC-grid} & 51.0 & 12.57 & 53.2 & 57.5 & 49.5  \\
\emph{all-grid} & 54.6 & 11.61 & 57.4 & 47.9 & 71.7  \\
\emph{\small{Geometry-Regressor}} & 52.9 & 11.05 & 54.2 & 48.2 & 62.0  \\
\cline{1-6}
\emph{NCs}& \textbf{62.8} & \textbf{9.54} & \textbf{65.4} & 65.5 & 65.4  \\

\hline
\end{tabular}
\end{center}
\vskip -2mm
\caption{Comparisons with competing methods using drawings from Cole et al.’s dataset and our newly collected dataset. IoU, F1, P, R are reported in percentages, CD is pixel distance.}
\vskip -5mm
\label{table:supp_exp}
\end{table}
\section*{5. Training Set Collection}
We created Amazon MTurk questionnaires to collect our training dataset. Each questionnaire had $35$ questions. $5$ of the questions were randomly chosen from a pool of $15$ sentinels. Each sentinel question showed eight line drawings along with renderings from a reference 3D model. One line drawing was created by an artist for the reference shape, and seven line drawings were created for different 3D models. The line drawings were presented to the participants in a random order.  Choosing one of the seven line drawings (or the option ``none of these line drawings are good'')  resulted in failing the sentinel. If a worker failed in one of these $5$ sentinels, then he/she was labeled as ``unreliable'' and the rest of his/her responses were ignored. A total of $4396$ participants took part in this user study to collect the training data. Among $4396$ participants, $657$ users ($15\%$) were labeled as ``unreliable''. Each participant was allowed to perform the questionnaire only once. 

\section*{6. Perceptual Evaluation}
We conducted an Amazon Mechanical Turk perceptual evaluation where we showed participants (a) a rendered shape from a viewpoint of interest along with two more views based on shifted camera azimuth by 30 degrees, (b) a pair of line drawings placed in a randomized left/right position: one line drawing was picked from our method, while the other came from \emph{pix2pixHD},  \emph{NC-geometry}, \emph{NC-image}, or \emph{AR-rtsc}. We asked participants to select the drawing that best conveyed the shown 3D  shape. Participants could pick one of four options:  left
drawing, right drawing, ``none
 of the drawings
conveyed the shape well'', or ``both” drawings conveyed the shape equally well''. The study included the $12$ shapes ($2$ viewpoints each)
 from both Cole \etal's and our new collected test dataset ($44$ shapes, two viewpoints each). Thus, there were total $112$ test cases, each involving the above-mentioned $4$ comparisons of techniques ($448$ total comparisons).

Each questionnaire was released via the MTurk platform. It contained $15$ unique questions, each asking for one comparison. Then these $15$ questions were repeated in the questionnaire in a random order. In these repeated questions, the order of compared line drawings was flipped. If a worker gave more than $7$ inconsistent answers for the repeated questions, then he/she was marked as ``unreliable''. Each participant was allowed to perform the questionnaire only once. A total of $225$ participants took part in the study. Among $225$ participants, $38$ workers were marked as ``unreliable''. For each of the $448$ comparisons, we gathered  consistent answers from $3$ different users. The results are shown in Figure 7 of the main text.

\end{document}